# ScenePilot-4K: A Large-Scale First-Person Dataset and Benchmark for Vision-Language Models in Autonomous Driving

Yujin Wang*
larswang@tongji.edu.cn
Tongji University
Shanghai, China

Yutong Zheng*
2052235@tongji.edu.cn
Tongji University
Shanghai, China

Wenxian Fan
2151472@tongji.edu.cn
Tongji University
Shanghai, China

Tianyi Wang
bonny.wang@utexas.edu
UT Austin
Austin, TX, USA

Hongqing Chu
chuhongqing@tongji.edu.cn
Tongji University
Shanghai, China

Li Zhang
lizhangfd@fudan.edu.cn
Fudan University
Shanghai, China

Bingzhao Gao†
gaobz@tongji.edu.cn
Tongji University
Shanghai, China

Daxin Tian
dtian@buaa.edu.cn
Beihang University
Beijing, China

Jianqiang Wang
wjqlws@tsinghua.edu.cn
Tsinghua University
Beijing, China

Hong Chen
chenhong2019@tongji.edu.cn
Tongji University
Shanghai, China

## Abstract

In this paper, we introduce ScenePilot-4K, a large-scale first-person dataset for safety-aware vision-language learning and evaluation in autonomous driving. Built from public online driving videos, ScenePilot-4K contains 3,847 hours of video and 27.7M front-view frames spanning 63 countries/regions and 1,210 cities. It jointly provides scene-level natural-language descriptions, risk assessment labels, key-participant annotations, ego trajectories, and camera parameters through a unified multi-stage annotation pipeline. Building on this dataset, we establish ScenePilot-Bench, a standardized benchmark that evaluates vision-language models along four complementary axes: scene understanding, spatial perception, motion planning, and GPT-based semantic alignment. The benchmark includes fine-grained metrics and geographic generalization settings that expose model robustness under cross-region and cross-traffic domain shifts. Baseline results on representative open-source and proprietary vision-language models show that current models remain competitive in high-level scene semantics but still exhibit substantial limitations in geometry-aware perception and planning-oriented reasoning. Beyond the released dataset itself, the proposed annotation pipeline serves as a reusable and extensible recipe for scalable dataset construction from public Internet driving videos. The codes and supplementary materials are available at: https://github.com/yjwangtj/ScenePilot-4K, with the dataset available at https://huggingface.co/datasets/larswangtj/ScenePilot-4K.

*Both authors contributed equally to this work.
†Corresponding author.

## 1 Introduction

Vision-language models (VLMs) recently show strong capabilities in multimodal understanding and reasoning, which motivates growing interest in their use for autonomous driving. Compared with conventional perception modules, VLMs provide a unified interface for scene description, risk reasoning, participant identification, and decision-related interpretation. However, autonomous driving imposes requirements that differ substantially from those of general-domain multimodal evaluation. In realistic driving scenarios, a useful model should not only describe what is visible, but also preserve ego-centric spatial relations, reason about safety-critical participants, and support planning-oriented interpretation under diverse road environments and traffic conventions.

Despite this demand, existing resources for driving-oriented VLM research remain fragmented. Large-scale autonomous driving datasets provide rich perception annotations, but they often lack language-centered supervision and unified evaluation for multimodal reasoning. Conversely, recent VLM-oriented driving benchmarks mainly focus on question answering or narrow task settings, and seldom jointly assess scene understanding, spatial grounding, and motion planning within a single evaluation framework. This gap makes it difficult to systematically measure how well modern VLMs support driving-oriented understanding in realistic and geographically diverse environments.

To address this problem, we introduce ScenePilot-4K, a large-scale first-person dataset for vision-language learning and evaluation in autonomous driving. It contains 3,847 hours of video and 27.7M front-view frames, and covers 63 countries/regions and 1,210 cities, with all original videos collected from YouTube and Bilibili. Through a unified multi-stage annotation pipeline, it jointly provides scene-level natural-language descriptions, risk assessment labels, key-participant annotations, ego trajectories, and camera parameters. This unified annotation design connects semantic, spatial, and planning-related supervision within the same clip-level data structure, thereby supporting more comprehensive evaluation of driving-oriented multimodal intelligence.

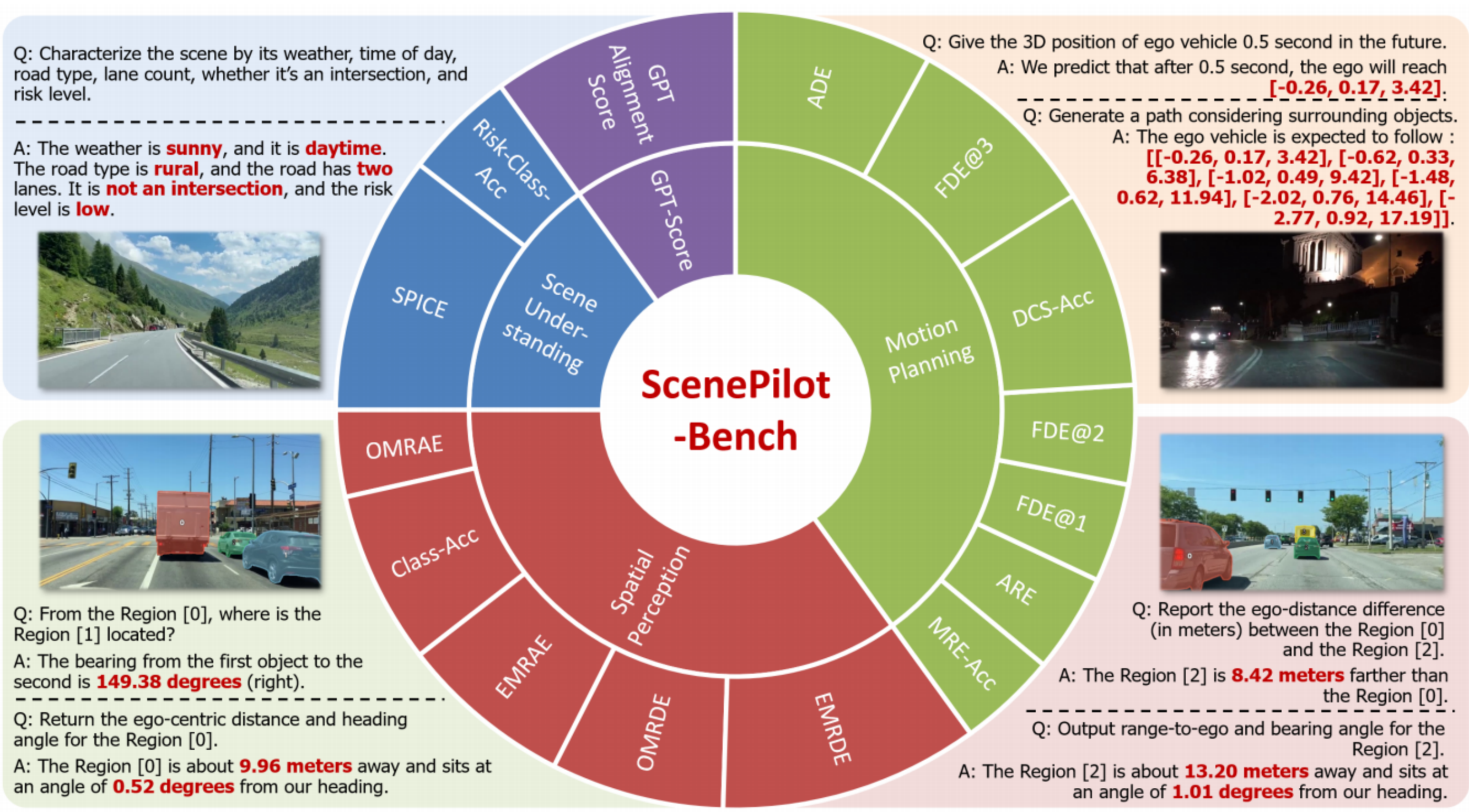

**Figure 1: The overall structure of ScenePilot-Bench benchmark, which emphasizes four critical metrics for VLM evaluation in autonomous driving: Scene Understanding, Spatial Perception, Motion Planning, and GPT-Score.**

Building on ScenePilot-4K, we establish ScenePilot-Bench, a standardized benchmark for evaluating VLMs in first-person autonomous driving scenarios, as illustrated in Figure 1. The benchmark organizes evaluation along four complementary axes: scene understanding, spatial perception, motion planning, and GPT-Score. Rather than measuring only semantic plausibility, ScenePilot-Bench is designed to assess driving-relevant abilities such as risk-sensitive reasoning, object-level spatial grounding, and trajectory-oriented prediction. In addition, we introduce geographic generalization settings to examine model robustness across regions with different traffic conventions and visual environments, which is critical for real-world deployment.

To facilitate reproducible comparison, we provide baseline results on representative VLMs. These results show a consistent pattern: current models show profound capabilities in high-level scene semantics, but still lack in spatial perception, embodied reasoning, and cross-region transfer. Such findings demonstrate that large-scale multimodal evaluation for autonomous driving should move beyond generic captioning or narrow visual question answering(VQA) settings toward more integrated assessment of semantics, spatial grounding, and motion-related reasoning. Therefore, ScenePilot-4K and ScenePilot-Bench are proposed to fill this gap and provide a comprehensive framework for evaluating and fine-tuning VLMs in safety-critical autonomous driving contexts.

Our contributions are summarized as follows: First, we introduce ScenePilot-4K, a large-scale and geographically diverse first-person driving dataset with unified multi-task annotations, together with a reusable annotation pipeline that supports scalable dataset extension. Second, built upon this dataset, we establish ScenePilot-Bench, a four-axis benchmark with extensible evaluation protocols for customized geographic generalization studies. Third, we provide baseline benchmarking and geographic generalization results over representative VLMs, which offer empirical evidence for current capability gaps and highlight open challenges for driving-oriented multimodal reasoning.

## 2 Related Work

Dedicated driving datasets are essential for improving scene understanding, spatial perception, and decision-related reasoning in autonomous driving. Early benchmarks, such as KITTI [11] and Cityscapes [7], focus on core perception and urban semantics, while larger-scale resources, including Waymo Open Dataset [20] and BDD100K [28], extend coverage to complex traffic scenes and multi-task video annotations. Language-grounded and risk-aware driving data further emerge in grounding and command-following settings [8, 16] and in risk-centric datasets such as DRAMA and DRAMA-ROLISP [9, 17]. For motion forecasting and planning, datasets such as nuScenes [3], nuPlan [4], and Argoverse 2 [24] support trajectory prediction with multi-agent dynamics and map context. More recent efforts, such as OpenDV-2K [27], further highlight the importance of scalable and diverse first-person driving data.

Systematic benchmarks are equally important for evaluating how vision-language models can be used in autonomous driving. Existing studies mainly follow two paradigms: open-loop and

**Table 1: Comparison among representative baseline datasets and ScenePilot-4K dataset in terms of scale and annotation.**

| Dataset | Duration | Frames | Countries & Regions | Cities | Scene Desc. | Risk Assess. | Key Partic. | Ego Traj. | Cam. Param. |
|---|---|---|---|---|---|---|---|---|---|
| KITTI [11] | 1.4h | 15K | 1 | 1 | × | × | ✓ | ✓ | ✓ |
| nuScenes [3] | 5.5h | 214K | 2 | 2 | × | × | ✓ | ✓ | ✓ |
| Waymo Open Perception [20] | 11h | 390K | 1 | 3 | × | × | ✓ | ✓ | ✓ |
| Waymo Open Motion [10] | 574h | 20.7M | 1 | 6 | × | × | ✓ | ✓ | ✓ |
| BDD100K [28] | 1,000h | 318K | 1 | 2 | ✓ | × | ✓ | ✓ | × |
| Cityscapes [7] | 0.5h | 25K | 3 | 50 | × | × | × | ✓ | ✓ |
| Argoverse 2 [24] | 4.2h | 300K | 1 | 6 | ✓ | × | ✓ | ✓ | ✓ |
| ApolloScape [23] | 103min | 12K | 1 | 1 | ✓ | × | ✓ | ✓ | ✓ |
| nuPlan [4] | 1282h | 62.5M | 2 | 4 | ✓ | × | ✓ | ✓ | ✓ |
| ONCE [18] | approx. 144h | approx. 1M | 1 | - | ✓ | × | ✓ | ✓ | ✓ |
| Lyft Level-5 [12] | approx. 1,000h | approx. 62.5M | 1 | 1 | ✓ | × | ✓ | ✓ | ✓ |
| Talk2Car [8] | 4.7h | 9.2K | 2 | 2 | × | × | ✓ | ✓ | ✓ |
| OpenDV-2K [27] | 2,059h | 65.1M | ≥40 | ≥244 | ✓ | × | × | × | × |
| **ScenePilot-4K (ours)** | **3,847h** | **27.7M*** | **63****| **1210**** | ✓ | ✓ | ✓ | ✓ | ✓ |

* Frame sampling rate is 2 Hz. ** Obtained based on video filenames.

closed-loop. Open-loop benchmarks assess perception and reasoning quality without directly controlling the vehicle, as represented by DriveLM [19], DriveBench [26], and object- or reasoning-centric benchmarks such as NuPrompt and STRIDE-QA [14, 25]. Closed-loop evaluation instead measures the downstream control effect of model outputs, as exemplified by BENCH2ADVLM [29]. However, current VLMs more often serve as auxiliary modules for semantic reasoning, spatial grounding, and planning separately. Existing resources therefore still lack a unified first-person dataset and benchmark that jointly support language-centered scene understanding, geometry-aware spatial perception, and planning-oriented evaluation. This gap motivates ScenePilot-4K and its accompanying benchmark, ScenePilot-Bench.

## 3 ScenePilot-4K Dataset

ScenePilot-4K is a large-scale first-person dataset that supports unified vision-language evaluation in autonomous driving. It contains 3,847 hours of driving videos and 27.7M front-view frames, spanning 63 countries/regions and 1,210 cities, as shown in Table 1 and Figure 2. Compared with prior driving datasets, this scale and geographic coverage better capture cross-region variation in road infrastructure, traffic behaviors, and driving conventions (e.g., left- vs. right-hand traffic), which is important for training and evaluating VLMs with real-world robustness.

A key property of ScenePilot-4K is its unified annotation richness. Each video clip consisting of 10 frames is associated with scene-level natural-language description, risk assessment, key-participant annotation, ego trajectory, and camera parameters, enabling joint evaluation of scene understanding, spatial perception, and motion planning. ScenePilot-4K also maintains broad diversity in scene composition and driving context. The dataset spans different weather conditions, day and night scenes, multiple road types, intersection and non-intersection layouts, different lane-count settings, and both right-hand and left-hand traffic systems. Moreover, it preserves a long-tailed risk distribution, which makes it suitable not only for large-scale training but also for robustness analysis in safety-critical and geographically shifted driving scenarios. This diversity is important not only for large-scale model training, but also for diagnosing whether driving-oriented VLMs remain robust under geographic changes.

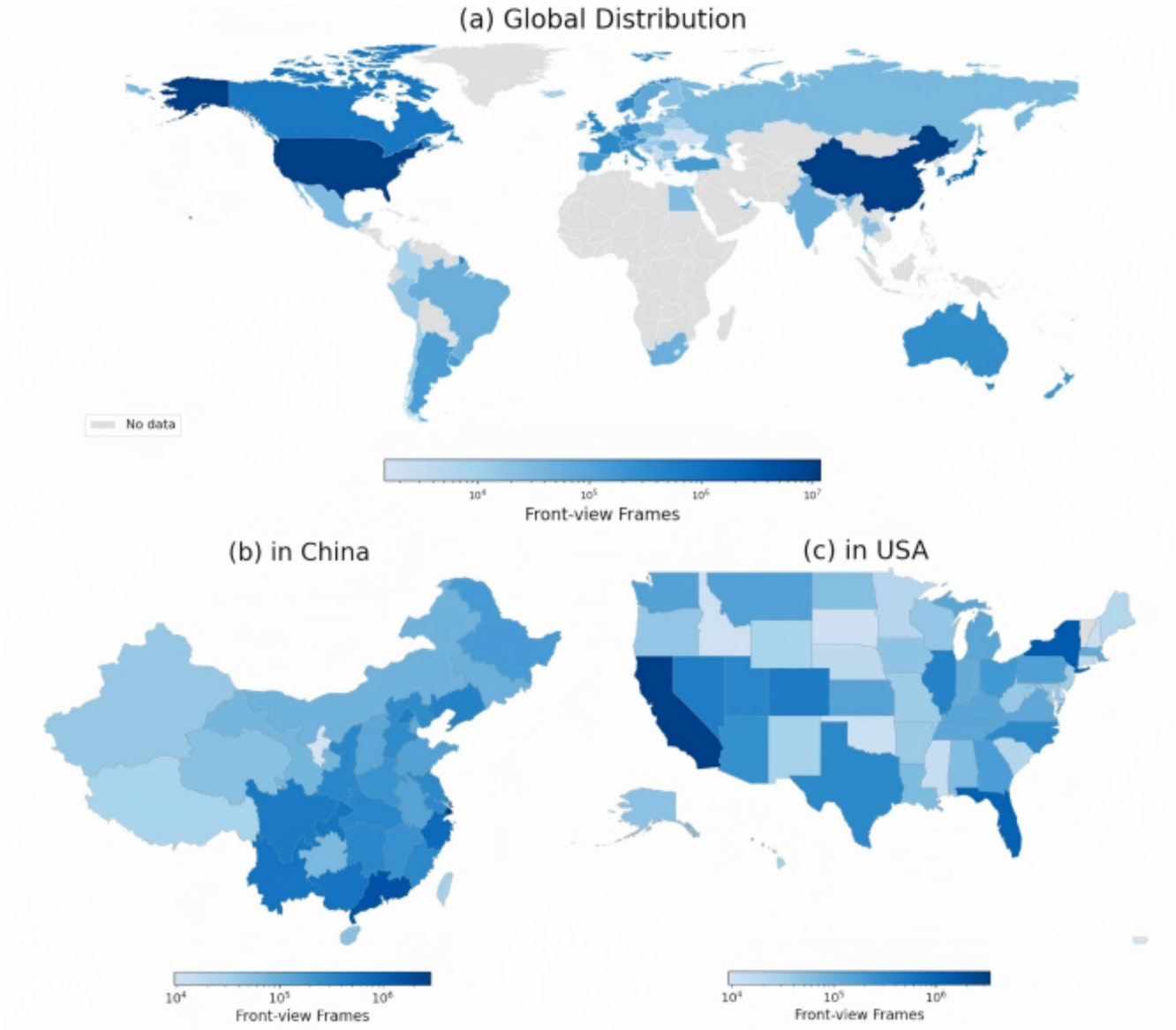

**Figure 2: The geographic distribution of ScenePilot-4K dataset. (a) The dataset covers the majority of developed countries as well as countries and regions with relatively well-established infrastructure. (b) The dataset covers all 34 provincial administrative regions in China, with a focus on the eastern coastal areas. (c) The dataset also covers all 50 states in the United States, with a focus on CA, FL, and NY.**

Raw videos are uniformly sampled at 2 FPS and segmented into 5-second clips with 10 frames for subsequent multi-task annotation. We adopt a unified automatic annotation pipeline to produce semantic, participant-level, and geometric labels for each clip.

For semantic annotation, we use Qwen2-VL-72B-Instruct [22] to generate a scene-level natural-language description and a risk level for each clip. In each clip, we use the 4th frame as a representative key frame for semantic querying. For participant annotation, we apply YOLO11s [15] with class-specific confidence thresholds to detect vehicles, trucks, bicycles, motorcycles, and pedestrians. For each detected participant, we record its class label and bounding box coordinates. For geometric annotation, we use the pre-trained VGGT model [21] to estimate camera intrinsics, extrinsics, and per-frame ego trajectory. Based on these outputs, we further perform lightweight geometric post-processing to lift 2D evidence into 3D space, recover metric scale, extract foregrounds for key participants, and derive ego-centric distances, azimuths, and inter-agent proximities. Based on these annotations, we generate over 400 million VQA pairs for downstream training and evaluation.

To assess the annotation quality, we also adopt cross-dataset validation on nuScenes [3] for trajectory annotation and on STRIDE-QA [14] for spatial perception. Failed annotations are erased rather than corrected, which ensures that the released annotations maintain a high level of reliability for training and evaluation. This conservative quality control strategy prioritizes annotation precision and benchmark reliability over raw annotation coverage. The details of the whole pipeline are available in supplementary materials.

ScenePilot-4K is constructed by extending OpenDV-2K [27]. While the original OpenDV-2K videos are sourced from YouTube, the newly added videos are collected from both YouTube and Bilibili. During data expansion, we follow the original distribution of OpenDV-2K as a reference and further supplement countries and regions that are underrepresented in the original collection. To improve data quality, we perform video-level deduplication and filter out low-quality content, especially clips with temporally discontinuous frames or severe overlaid-text occlusions.

For public release, we provide the codes for data processing and annotation, as well as the derived annotations. Specifically, HuggingFace hosts the complete annotation results together with a table of source-video links, while GitHub provides the codes. Raw source videos and extracted frames could be accessed from the original platforms subject to their respective terms of use. This release protocol supports reproducible academic research while avoiding direct redistribution of third-party source media.

Importantly, ScenePilot-4K is not intended to be only a fixed released dataset, but also a reusable and extensible data construction framework. Because our annotation pipeline starts from publicly available Internet driving videos and automatically derives semantic, participant-level, and geometric annotations under a unified schema, researchers can readily adapt it to build customized datasets for new regions, traffic conventions, or downstream tasks. This design substantially improves the scalability of ScenePilot-4K and makes the dataset construction process itself reusable by the community.

## 4 ScenePilot-Bench Benchmark

Built upon ScenePilot-4K, ScenePilot-Bench serves as a standardized open-loop evaluation protocol for vision-language models in autonomous driving. The benchmark assesses four complementary capabilities: scene understanding, spatial perception, motion planning, and GPT-based semantic judgment. Rather than focusing only on caption fluency or generic VQA, it evaluates whether model outputs remain consistent with semantics, spatial grounding, and short-horizon planning in realistic traffic scenes.

**Table 2: Summary of evaluation metrics in ScenePilot-Bench.**

| Task | Metric | Description |
|---|---|---|
| Scene Understanding | SPICE | Measures semantic quality of generated scene descriptions. |
| | Risk-Class-Acc | Accuracy of predicted scene risk level. |
| Spatial Perception | Class-Acc | Accuracy of key participant category prediction. |
| | EMRDE / EMRAE | Ego-to-object relative distance / azimuth error. |
| | OMRDE / OMRAE | Object-to-object relative distance / azimuth error. |
| Motion Planning | DCS-Acc | Accuracy of predicted discrete driving behavior or command. |
| | MRE-Acc | Accuracy under motion-related regression tolerance. |
| | ARE | Angular relative error of the predicted trajectory. |
| | ADE | Average displacement error of the predicted trajectory. |
| | FDE@k | Final displacement error at future k seconds. |
| GPT-Score | GPT-Score | LLM-based holistic judgment of response quality. |

Specifically, scene understanding measures scene description quality and risk level prediction. Spatial perception evaluates key participant identification together with ego-relative and inter-object geometric reasoning. Motion planning examines action-related reasoning and short-horizon trajectory prediction. GPT-Score provides an additional holistic judgment of response quality. Table 2 summarizes the corresponding evaluation tasks and metrics. The overall score is not computed from raw metrics directly; instead, each metric is first normalized and aggregated within its corresponding axis, and the final benchmark score is then obtained by combining the four axis-level scores. This factorized design makes it possible to distinguish whether model gains come from semantic fluency, spatial grounding, or planning consistency, which are often conflated in caption-only or VQA-only evaluation. Detailed normalization and weighting strategies are provided in supplementary materials.

In addition to in-domain evaluation, ScenePilot-Bench explicitly emphasizes geographic generalization. Autonomous driving models must operate under diverse road layouts, traffic conventions, and visual environments, and should therefore be evaluated beyond a single-country distribution. We accordingly define generalization settings across different regions and traffic systems, including both country-level hold-out evaluation and right-to-left traffic adaptation. These settings make the benchmark more realistic and more diagnostic than in-domain testing alone. More importantly, these geographic generalization settings are intended as representative use cases rather than exhaustive evaluation protocols, and users can flexibly define customized cross-region, cross-country, or cross-traffic experiments based on their own research goals.

**Table 3: Baseline results on ScenePilot-Bench.**

| Model | Scene Understanding | | | Spatial Perception | | | | | |
|---|---|---|---|---|---|---|---|---|---|
| | SPICE | Risk-Class-Acc | Total | Class-Acc | EMRDE | EMRAE | OMRDE | OMRAE | Total |
| GPT-4o[13] | 92.93 | 74.82 | 87.50 | 91.93 | 49.57 | 22.51 | 26.23 | 22.23 | 45.23 |
| GPT-5[2] | 92.18 | 71.64 | 86.01 | 91.83 | 57.58 | 25.83 | 19.37 | 23.92 | 47.07 |
| Gemini-2.5-flash[6] | 91.70 | 69.44 | 85.02 | **92.55** | 54.39 | 22.70 | 22.24 | 24.63 | 46.28 |
| Qwen3-VL-235B[1] | 93.58 | 77.23 | 88.68 | 91.64 | 57.92 | 14.99 | 21.87 | 21.61 | 45.24 |
| ReasonDrive-7B[5] | 85.31 | 33.84 | 69.86 | 75.12 | 74.10 | 7.27 | 10.00 | 52.22 | 45.92 |
| ReasonDrive-7B + ScenePilot-4K FT | 85.45 | 37.06 | 70.93 | 89.27 | 71.81 | 26.83 | 31.44 | **71.28** | 58.18 |
| ReasonDrive-7B-CoT[5] | 85.51 | 34.07 | 70.08 | 88.55 | 53.88 | 7.36 | 8.51 | 36.87 | 40.73 |
| ReasonDrive-7B-CoT + ScenePilot-4K FT | 87.90 | 56.91 | 78.60 | 90.18 | 68.81 | 22.16 | 23.93 | 51.51 | 53.05 |
| Qwen2-VL-2B + ScenePilot-4K FT | 93.64 | **80.57** | **89.72** | 85.74 | 87.76 | 73.67 | 48.41 | 50.75 | 72.97 |
| Qwen2.5-VL-3B + ScenePilot-4K FT | **93.65** | 78.67 | 89.15 | 86.19 | **87.98** | **73.69** | **52.58** | 55.78 | **74.46** |

| Model | Motion Planning | | | | | | | | GPT-Score | Overall |
|---|---|---|---|---|---|---|---|---|---|---|
| | DCS-Acc | MRE-Acc | ARE | ADE | FDE@1 | FDE@2 | FDE@3 | Total | | |
| GPT-4o[13] | 15.79 | 82.16 | 85.07 | 58.68 | 85.79 | 42.71 | 24.15 | 49.30 | 22.56 | 50.93 |
| GPT-5[2] | 22.39 | 81.32 | 79.59 | 57.53 | 85.25 | 44.88 | 28.20 | 50.73 | 36.51 | 53.32 |
| Gemini-2.5-flash[6] | 19.59 | 76.73 | 79.89 | 43.77 | 76.40 | 34.10 | 19.63 | 43.31 | 33.64 | 49.64 |
| Qwen3-VL-235B[1] | 22.80 | 43.83 | 83.05 | 12.69 | 15.70 | 8.67 | 5.55 | 23.33 | 34.22 | 41.89 |
| ReasonDrive-7B[5] | 14.38 | 29.45 | 61.23 | 13.67 | 23.31 | 11.25 | 9.09 | 19.95 | 15.70 | 36.10 |
| ReasonDrive-7B + ScenePilot-4K FT | 10.50 | 95.25 | 93.01 | **98.26** | 99.51 | 97.48 | **82.95** | 76.87 | 27.60 | 64.51 |
| ReasonDrive-7B-CoT[5] | 14.12 | 0.49 | 47.55 | 0.00 | 0.00 | 0.00 | 0.00 | 7.63 | 17.90 | 29.61 |
| ReasonDrive-7B-CoT + ScenePilot-4K FT | 11.14 | **99.55** | **95.18** | 98.21 | **99.54** | **97.52** | 82.70 | **77.59** | 24.72 | 63.86 |
| Qwen2-VL-2B + ScenePilot-4K FT | 28.46 | 79.47 | 75.07 | 59.31 | 82.72 | 47.78 | 29.16 | 51.89 | 52.53 | 65.01 |
| Qwen2.5-VL-3B + ScenePilot-4K FT | **29.52** | 88.20 | 70.09 | 53.07 | 79.94 | 43.64 | 32.67 | 51.24 | **54.43** | **65.37** |

# 5 Experiments

## 5.1 Baseline Results on ScenePilot-Bench

We evaluate representative commercial, open-source, and driving-specialized VLMs on ScenePilot-Bench. ScenePilot-4K is partitioned into training, validation, and test splits with balanced coverage over weather conditions, road types, lane configurations, intersection attributes, risk levels, traffic density, and geographic regions. To avoid data leakage, all training and test samples are drawn from disjoint video splits. For baseline fine-tuning, we sample 200,000 training instances and 100,000 test instances from the corresponding splits. Additional implementation details are provided in the supplementary material.

Table 3 reports three baseline groups: VLMs without driving-specific fine-tuning, driving-specialized ReasonDrive [5] variants, and Qwen backbones fine-tuned on ScenePilot-4K. The results reveal a consistent pattern. General-purpose VLMs remain strong in scene-level semantics, as reflected by high SPICE scores and competitive risk understanding, but their performance degrades substantially when evaluation requires ego-centric spatial grounding and planning-oriented reasoning. This gap suggests that generic multimodal pretraining transfers well to coarse semantic understanding, yet does not reliably induce the geometric calibration and reference-frame consistency required in autonomous driving. Driving-specialized models, such as ReasonDrive-7B [5], improve several driving-oriented metrics after fine-tuning on ScenePilot-4K, especially in the evaluation of motion planning. Fine-tuned Qwen backbones achieve the most balanced performance across scene understanding, spatial perception, and motion planning. Rather than indicating a benchmark-specific optimization advantage, this result shows that ScenePilot-4K provides effective supervision for jointly improving semantic reasoning, spatial grounding, and short-horizon planning. Overall, Table 3 demonstrates that ScenePilot-Bench serves as a diagnostic benchmark that clearly separates strengths in high-level semantics from weaknesses in embodied perception and planning.

## 5.2 Geographic Generalization

We further evaluate cross-region robustness under geographic domain shift. In the country-level hold-out setting, we train the Qwen2.5-VL-3B baseline on 200,000 samples from China and test it on four unseen regions: Europe (EU), Japan/Korea (JP/KR), the United States (US), and other countries (OTH), each with 100,000 samples. We also evaluate a right-to-left traffic adaptation setting, where the model is trained only on right-hand-traffic countries and tested on left-hand-traffic regions. Table 4 summarizes the results.

The results show that scene-level semantics transfer more robustly than planning-oriented reasoning under geographic shift. SPICE remains consistently high across regions, while risk understanding shows moderate variation. Spatial perception remains broadly functional, but the geometric metrics are more sensitive to regional differences in road layout, visual depth cues, and traffic organization. Motion planning exhibits the largest degradation, especially in discrete driving behavior prediction, indicating that high-level action reasoning is more dependent on local traffic conventions than short-horizon geometric prediction. The right-to-left adaptation results show a similar trend: scene understanding and spatial perception remain relatively stable, whereas motion planning is more strongly affected by traffic-direction shifts. These findings suggest that ScenePilot-Bench provides a practically relevant robustness test beyond in-domain evaluation. This also suggests that future driving-oriented VLMs should be evaluated not only by in-domain accuracy, but also by their ability to preserve spatial and planning consistency under realistic cross-region distribution shifts.

**Table 4: The experiment on geographic generalization of VLMs.**

| Model | Scene Understanding | | | Spatial Perception | | | | | |
|---|---|---|---|---|---|---|---|---|---|
| | SPICE | Risk-Class-Acc | Total | Class-Acc | EMRDE | EMRAE | OMRDE | OMRAE | Total |
| SP-2.5-3B-CN (EU) | 92.57 | 75.26 | 87.37 | 83.44 | 91.53 | 67.33 | 54.37 | 54.38 | 73.92 |
| SP-2.5-3B-CN (JP/KR) | 92.48 | 71.01 | 86.04 | 83.84 | 89.36 | 68.50 | 53.91 | 56.60 | 73.72 |
| SP-2.5-3B-CN (US) | 92.74 | 74.75 | 87.34 | 83.18 | 90.33 | 69.04 | 56.72 | 56.76 | 74.56 |
| SP-2.5-3B-CN (OTH) | 91.99 | 70.14 | 85.44 | 83.09 | 92.35 | 71.16 | 57.74 | 56.68 | 75.77 |
| SP-2.5-3B-R (L) | 93.72 | 78.79 | 89.24 | 90.51 | 85.21 | 74.17 | 54.51 | 57.10 | 75.11 |

| Model | Motion Planning | | | | | | | | GPT-Score | Overall |
|---|---|---|---|---|---|---|---|---|---|---|
| | DCS-Acc | MRE-Acc | ARE | ADE | FDE@1 | FDE@2 | FDE@3 | Total | | |
| SP-2.5-3B-CN (EU) | 12.72 | 65.11 | 50.20 | 75.33 | 74.40 | 69.76 | 70.24 | 57.60 | 54.54 | 67.47 |
| SP-2.5-3B-CN (JP/KR) | 12.80 | 68.83 | 53.03 | 75.37 | 70.61 | 67.03 | 69.18 | 57.42 | 54.96 | 67.17 |
| SP-2.5-3B-CN (US) | 11.69 | 64.03 | 51.27 | 81.18 | 72.53 | 69.79 | 76.78 | 59.69 | 54.14 | 68.49 |
| SP-2.5-3B-CN (OTH) | 12.73 | 66.95 | 52.17 | 84.67 | 81.50 | 78.11 | 80.30 | 63.41 | 56.22 | 70.32 |
| SP-2.5-3B-R (L) | 10.95 | 54.68 | 47.68 | 78.95 | 80.77 | 74.95 | 73.70 | 58.53 | 54.61 | 68.55 |

## 6 Conclusion

This paper presents ScenePilot-4K, a large-scale first-person driving dataset for vision-language learning and evaluation in autonomous driving. ScenePilot-4K contains 3,847 hours of driving videos and 27.7M front-view frames across 63 countries/regions and 1,210 cities, and provides unified annotations for scene description, risk assessment, key participant annotation, ego trajectory, and camera parameters. These properties make it a geographically diverse and annotation-rich resource for autonomous driving.

Built upon ScenePilot-4K, ScenePilot-Bench provides a standardized open-loop benchmark that evaluates vision-language models from four complementary perspectives: scene understanding, spatial perception, motion planning, and GPT-based semantic judgment. In addition to in-domain evaluation, the benchmark includes geographic generalization settings that probe robustness under cross-region and cross-traffic domain shifts. Together, the dataset and benchmark support more comprehensive analysis than prior resources that focus only on perception labels, narrow language tasks, or isolated planning metrics. Beyond the released benchmark itself, we expect the proposed annotation pipeline and evaluation protocol to serve as extensible tools for constructing customized driving datasets and defining new robustness tests under diverse geographic settings.

Baseline results on representative commercial, open-source, and driving-specialized vision-language models reveal a consistent pattern: current models remain relatively strong in high-level scene semantics, but still show clear limitations in geometry-aware perception, planning-oriented reasoning, and robustness under geographic shift. These findings indicate that driving-oriented vision-language evaluation requires more than generic captioning or VQA.

Despite these strengths, the current release still has several limitations. First, ScenePilot-Bench remains an open-loop benchmark, and therefore does not directly measure the downstream control effect of model outputs in a closed-loop driving system. Second, although we adopt cross-dataset validation and conservative filtering to improve annotation reliability, part of the annotations are still automatically derived from VLM and geometric reconstruction pipelines, which may introduce residual noise in difficult scenes. Third, the current geographic generalization settings are representative rather than exhaustive, and broader protocols across camera setups, weather patterns, and regional distributions remain to be explored. We view more human verification, broader cross-domain evaluation, and future closed-loop integration as important directions for extending this benchmark.

We expect ScenePilot-4K and ScenePilot-Bench to serve as useful open resources for future research on multimodal driving understanding, spatial grounding, and safety-aware evaluation. By releasing annotations, metadata, split files, source-video indices, and reproduction code, we hope to support more transparent, reproducible, and geographically aware research on multimodal driving intelligence.

## Acknowledgments

This research is supported by National Key R&D Program of China (2023YFB2504400), the National Nature Science Foundation of China (No. 62373289), Shanghai Municipal Science and Shanghai Automotive Industry Science and Technology Development Foundation (No.2407), and the Fundamental Research Funds for the Central Universities.

# Supplementary Material for ScenePilot-4K: A Large-Scale First-Person Dataset and Benchmark for Vision-Language Models in Autonomous Driving

Yujin Wang*
larswang@tongji.edu.cn
Tongji University
Shanghai, China

Yutong Zheng*
2052235@tongji.edu.cn
Tongji University
Shanghai, China

Wenxian Fan
2151472@tongji.edu.cn
Tongji University
Shanghai, China

Tianyi Wang
bonny.wang@utexas.edu
UT Austin
Austin, TX, USA

Hongqing Chu
chuhongqing@tongji.edu.cn
Tongji University
Shanghai, China

Li Zhang
lizhangfd@fudan.edu.cn
Fudan University
Shanghai, China

Bingzhao Gao†
gaobz@tongji.edu.cn
Tongji University
Shanghai, China

Daxin Tian
dtian@buaa.edu.cn
Beihang University
Beijing, China

Jianqiang Wang
wjqlws@tsinghua.edu.cn
Tsinghua University
Beijing, China

Hong Chen
chenhong2019@tongji.edu.cn
Tongji University
Shanghai, China

## 1 Details of ScenePilot-4K Dataset

**Scale.** ScenePilot-4K aggregates 3,847 hours of driving videos with 27.7M front-view frames, making it among the largest video corpora for autonomous driving. Although some planning datasets report tens of millions of frames, none of them pair scale with the multi-task, language-grounded supervision offered here.

**Geographic diversity.** The dataset spans 63 countries/regions and 1,210 cities, far exceeding the predominantly single-country coverage of most benchmarks (often 1–6 cities). Even compared with the most geographically varied prior work (≥40 countries and ≥244 cities), ScenePilot-4K increases country coverage by over 50% and city coverage by about an order of magnitude. This breadth naturally captures diverse road infrastructures, signage systems, traffic behaviors, and driving conventions (e.g., left- vs. right-hand traffic), improving out-of-distribution robustness.

**Annotation richness.** ScenePilot-4K is, to our knowledge, the only dataset that jointly provides all of the following on the same clips: (i) scene-level natural-language descriptions, (ii) risk assessment labels, (iii) key-participant identification, (iv) ego-trajectory, and (v) camera parameters. In contrast, prior datasets typically offer only perception labels (boxes/segments) or metadata-derived ego trajectories, and when language or risk information is available it is usually limited ("only labels/meta-data") and not aligned with trajectory or calibration information. This unified, multi-granularity supervision enables learning objectives that bridge vision-language understanding, risk assessing, agent saliency, and planning-aware prediction within a single benchmark.

As illustrated in Fig. 1, ScenePilot-4K covers diverse real-world conditions, with scenario statistics broadly matching everyday driving (clear-weather dominance and limited but non-negligible night-time coverage). The risk distribution is intentionally long-tailed, preserving a small yet essential fraction of high-risk clips to support stress-testing of perception and reasoning. Moreover, the dataset spans multiple countries/regions and includes both right- and left-hand traffic conventions, which facilitates out-of-distribution evaluation under cross-regional shifts.

*Both authors contributed equally to this work.
†Corresponding author.

## 2 The High-quality Annotation Pipeline

**Video Clipping and Frame Sampling.** Raw first-person driving videos are preprocessed by removing the first and last 180 seconds. Within the valid range, frames are uniformly sampled at 2 FPS. The sampled frames are segmented into non-overlapping clips of 5 seconds, corresponding to 10 frames per clip. This pipeline supports large-scale batch processing with resume functionality and consistent data structure.

**Scene Description and Risk Assessment.** We apply a cross-modal annotation module to generate both a semantic scene description and a risk level score for each video clip. This process is automated using Qwen2-VL-72B-Instruct VLM.

For each video clip of $G = 10$ consecutive frames, only the 4th frame is selected as a representative key frame. The annotated scene description is formatted as: *The weather is sunny, and it is daytime. The road type is urban and the road has four lanes. It is an intersection, and the risk level score is 4.*

**Key Participant Detection and Annotation.** We detect traffic participants using YOLO11s with class-specific confidence thresholds. Given significant variation in object scale and visual distinctiveness, we define a per-class confidence threshold $\tau_c$ for each category $c \in C$, where:

$$C = \{\text{vehicle}, \text{truck}, \text{bicycle}, \text{motorcycle}, \text{pedestrian}\} \tag{1}$$

We set $\tau_{\text{vehicle}} = 0.5$, $\tau_{\text{truck}} = 0.6$, $\tau_{\text{bicycle}} = 0.4$, $\tau_{\text{motorcycle}} = 0.55$, $\tau_{\text{pedestrian}} = 0.55$, balancing recall and false positives by allowing frequent, reliable classes to use lower thresholds and applying stricter filtering to rarer or more ambiguous ones.

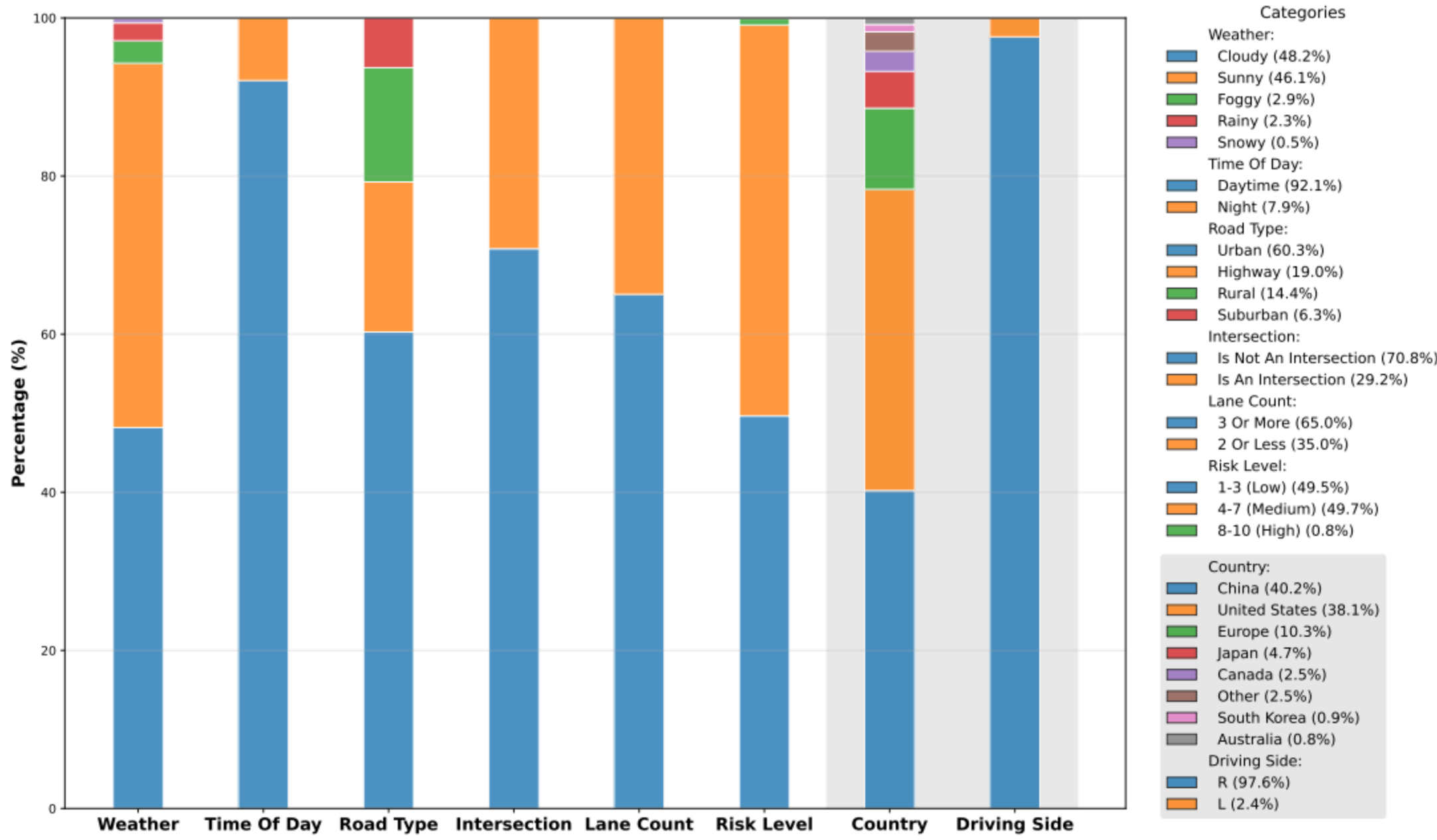

**Figure 1: ScenePilot-4K Dataset Statistics: Distribution of Scene Attributes and Driving Countries. This figure summarizes the overall distribution of scene attributes and geographic coverage in the ScenePilot-4K dataset.**

Besides, we store a JSON record per object (ID, label, normalized box), which can be directly combined with camera intrinsics and extrinsics.

**Camera Calibration and Ego-Trajectory Annotation.** We estimate camera intrinsics, extrinsics, and per-frame ego-trajectory using the pre-trained VGGT model. This enables frame-level geometric reasoning and future trajectory labeling without external sensors.

Given the video clip of 10 consecutive frames $\{I_t\}_{t=1}^{10}$, we apply VGGT to jointly infer camera parameters. The model outputs per-frame pose encodings, which are converted to camera matrices using calibrated functions.

Extrinsics are converted to camera-to-world matrices $\mathcal{T}_t^{c2w} \in \mathbb{R}^{4\times4}$ via SE(3) inversion. Matrices are adjusted to ensure positive translation values, yielding consistent world coordinate alignment.

We define the ego-trajectory as the camera centers $\mathbf{C}_t$ across frames, expressed in the first frame's world coordinate system. Each camera center is computed as:

$$\mathbf{C}_t = (-R_t^\top t_t) - (-R_1^\top t_1), \tag{2}$$

where $(R_1, t_1)$ are from frame $t = 1$. Small positional offsets are applied for stability. The result is a sequence of $T = 10$ world-frame positions:

$$\mathcal{T}_{\text{ego}} = \{\mathbf{C}_t\}_{t=1}^{10} \in \mathbb{R}^{10\times3}. \tag{3}$$

After obtaining per-frame camera intrinsics/extrinsics and ego trajectory with VGGT, we perform a lightweight geometric post-processing stage that: (i) lifts 2D evidence to 3D, (ii) recovers metric scale, (iii) extracts foregrounds for key participants, and (iv) reports robust ego-centric distances, azimuths, and inter-agent proximities suitable for VLM/VLA training and evaluation. The pipeline is fully automatic and runs per clip.

**3D lifting from monocular geometry.** For frame $t$, pixel $\mathbf{u} = (u, v)$, and depth $\hat{Z}_t(\mathbf{u})$,

$$\mathbf{p}_t^c(\mathbf{u}) = \hat{Z}_t(\mathbf{u}) K_t^{-1} \tilde{\mathbf{u}}, \quad \tilde{\mathbf{u}} = [u, v, 1]^\top, \tag{4}$$

where $\mathbf{p}_t^c(\mathbf{u}) \in \mathbb{R}^3$ is the back-projected 3D point in the camera frame (under VGGT units), $K_t \in \mathbb{R}^{3\times3}$ is the intrinsic matrix, and $\tilde{\mathbf{u}}$ is the homogeneous pixel.

Optionally, transform to the world frame by

$$\tilde{\mathbf{p}}_t^w(\mathbf{u}) = T_t^{c\to w} \left[\mathbf{p}_t^c(\mathbf{u})^\top, 1\right]^\top, \tag{5}$$

where $T_t^{c\to w} \in SE(3)$ is the camera-to-world transform and $\tilde{\mathbf{p}}_t^w(\mathbf{u}) = \left[(\mathbf{p}_t^w)^\top, 1\right]^\top$ is the homogeneous world point.

**Metric scale recovery (frame-level with object-level fallback).** To convert VGGT's arbitrary scale into metric distances, we first estimate a frame-level ground scale and then optionally fall back to class-specific object priors. We sample a grid $S$ in the lower half of the image, back-project each sample to a 3D point $p_{t,k}^c = (x_k, y_k, z_k)$ in the frame, and use robust statistics on the vertical coordinates $|y_k|$ to select ground inliers:

$$m = \operatorname{median}\left(\{|y_k|\}\right), \tag{6}$$

$$\text{MAD} = \operatorname{median}\left(\left||y_k| - m\right|\right), \tag{7}$$

$$\mathcal{G} = \{k : \left||y_k| - m\right| \le \kappa \cdot \text{MAD}\}, \tag{8}$$

where $\kappa > 0$ is a fixed threshold (we use $\kappa \approx 2.5$).

For each target $i$ in frame $t$, we then define a single working metric scale $s_{t,i}$ by combining the frame-level and object-level estimates

in a piecewise form:

$$s_{t,i} = \begin{cases} \dfrac{H_{\text{cam}}}{\text{median}(|y_k| : k \in \mathcal{G})}, & \text{if the frame-level estimate is reliable,} \\ \dfrac{H_{c(i)}}{(h_{\text{px}}/f_y)\tilde{z}}, & \text{otherwise,} \end{cases} \tag{9}$$

where $H_{\text{cam}} > 0$ is the nominal camera height, $H_{c(i)}$ is the canonical physical height for class $c(i)$, $h_{\text{px}}$ is the bounding-box height in pixels, $f_y$ is the vertical focal length, and $\tilde{z}$ is the median foreground depth (in VGGT units) for object $i$. In practice, if $\mathcal{G}$ is empty or unstable, we approximate the numerator term in the first branch using $m$ instead of median $(|y_k| : k \in \mathcal{G})$. The resulting $s_{t,i}$ (meters per VGGT unit) is used to scale all subsequent ego-centric and inter-object distance measurements.

**Foreground extraction for key participants.** Given a detection bbox, we obtain a binary mask $M_{t,i} \subset \Omega$ via SAM, with the bbox as the prompt, and then refine it with light morphology:

$$M_{t,i} = \text{Close}\left(\text{Open}\left(\text{SAM}(\text{bbox}, I_t)\right)\right), \tag{10}$$

where $I_t \in \mathbb{R}^{H \times W \times 3}$ is the RGB image, and Open/Close are standard 3×3 morphological operations. For vehicles, we ignore a thin bottom strip during seeding to reduce road leakage; if SAM inference fails or the bbox is overly large, $M_{t,i}$ falls back to the rectangle. The set of 3D foreground points is

$$\mathcal{P}_{t,i} = \left\{ \mathbf{p}_t^c(\mathbf{u}) : \mathbf{u} \in M_{t,i} \right\}. \tag{11}$$

**Robust ego-centric distance and azimuth.** Define per-point radii $d_j = \left\| \mathbf{p}_{t,i,j}^c \right\|_2$ for $\mathbf{p}_{t,i,j}^c \in \mathcal{P}_{t,i}$. We report a robust distance using a small percentile:

$$\tilde{d}_{t,i} = \text{perc}_5\left(\{d_j\}\right), \quad D_{t,i} = \max\{0, s_{t,i}\tilde{d}_{t,i}\}, \tag{12}$$

where $\text{perc}_5$ is the 5th percentile operator, $\tilde{d}_{t,i}$ is in VGGT units, and $D_{t,i}$ is in meters. For directional context,

$$\bar{\mathbf{p}}_{t,i}^c = \frac{1}{|\mathcal{P}_{t,i}|} \sum_{\mathbf{p} \in \mathcal{P}_{t,i}} \mathbf{p}, \quad \theta_{t,i} = \text{atan2}(\bar{x}, \bar{z}), \tag{13}$$

where $\bar{\mathbf{p}}_{t,i}^c = (\bar{x}, \bar{y}, \bar{z})$ is the foreground centroid in the camera frame and $\theta_{t,i} \in \left[-\frac{\pi}{2}, \frac{\pi}{2}\right]$ is the left/right azimuth (radians) relative to the forward axis.

**Inter-agent proximity.** For targets $i$ and $j$ in the same frame,

$$\delta_{i \to j} = \text{perc}_5\left(\left\{ \min_{\mathbf{q} \in \mathcal{P}_{t,j}} \|\mathbf{p} - \mathbf{q}\|_2 : \mathbf{p} \in \mathcal{P}_{t,i} \right\}\right), \tag{14}$$

where $\delta_{i \to j}$ is a robust, directed separation in VGGT units. The symmetric metric proximity is

$$\Delta_t(i,j) = s_t^\star \cdot \min\{\delta_{i \to j}, \delta_{j \to i}\}, \quad s_t^\star = \frac{1}{2}\left(s_{t,i} + s_{t,j}\right), \tag{15}$$

where $\Delta_t(i,j)$ is in meters and $s_t^\star$ averages the two targets' scales.

**Output.** For each frame, we save for each detection $i$ : $D_{t,i}$,$\theta_{t,i}$, class label, bbox, and a *scale_info* record indicating whether $s_t^{\text{frame}}$ or $s_{t,i}^{\text{obj}}$ was used (with diagnostics such as inlier counts/MAD). We also persist masks and 3D points (VGGT units and metric via $s_{t,i}$) for reproducibility.

# 3 Details of ScenePilot-Bench Benchmark

## 3.1 Scene Understanding

*3.1.1 Refined SPICE.* SPICE (Semantic Scene Consistency) evaluates the semantic propositional consistency between model-generated and reference scene graphs. We adopt a refined SPICE with a lightweight, Python-based parser tailored for autonomous driving. It uses a rule-based approach with domain-specific heuristics to extract key traffic participants and attribute tuples, avoiding dependency parsing. Relation tuples are deliberately omitted to reduce noise and improve robustness.

Formally, given a generated scene description $c$ and a ground truth description $S$, they are both parsed into tuple sets $T(G(c))$ and $T(G(S))$. Each tuple includes a unary object tuple $(o)$ and an attribute tuple $(o, a)$. Precision $(P)$, Recall $(R)$, and the final SPICE score are defined as:

$$P(c,S) = \frac{|\,T(G(c)) \otimes T(G(S))\,|}{|\,T(G(c))\,|}, \tag{16}$$

$$R(c,S) = \frac{|\,T(G(c)) \otimes T(G(S))\,|}{|\,T(G(S))\,|}, \tag{17}$$

$$\text{SPICE}(c,S) = F_1(c,S) = \frac{2\,P(c,S) \cdot R(c,S)}{P(c,S) + R(c,S)}, \tag{18}$$

where $\otimes$ denotes tuple matching with stemming and synonym normalization. A higher SPICE indicates better global semantic consistency.

*3.1.2 Risk-Class-Acc: Risk Reasoning Accuracy.* We introduce Risk-Class-Acc to quantify VLMs' performance in risk classification tasks:

$$\text{Risk-Class-Acc} = \frac{1}{N} \cdot \sum_{i=1}^{N} I(r_i^{\text{pred}} = r_i^{\text{gt}}), \tag{19}$$

where $I(\cdot)$ denotes the indicator function, while $r_i^{\text{pred}}$ and $r_i^{\text{gt}}$ represent the predicted and ground-truth risk classes, respectively. The risk classes are divided into three categories: low, medium, and high. This metric reflects VLMs' ability to perform safety-critical reasoning at the semantic level.

## 3.2 Spatial Perception

This section assesses VLMs' spatial perception ability to detect and interpret key traffic participants, such as vehicles, trucks, pedestrians, bicycles and motorcycles, under complex traffic scenarios. It consists of two modules: Object Classification and Spatial Reasoning.

*3.2.1 Object Classification.* In addition to recognition accuracy, VLMs are also required to possess precise category classification capability. We compute the overall accuracy directly on the entire test set as the core evaluation metric:

$$\text{Class-Acc} = \frac{1}{N} \sum_{i=1}^{N} I(c_i^{\text{pred}} = c_i^{\text{gt}}), \tag{20}$$

where $N$ is the total number of samples across all traffic participant categories, namely vehicle, truck, pedestrian, bicycle and motorcycle. This metric comprehensively reflects the overall recognition ability of VLMs across all categories.

*3.2.2 Spatial Reasoning.* To evaluate spatial reasoning, we measure relative positional and angular errors between the ego vehicle and traffic participants, as well as among participants.

Ego-centric metrics, which include Mean Relative Distance Error to Ego (EMRDE) and Mean Relative Angle Error to Ego (EMRAE), evaluate the relative position of each traffic participant with respect to the ego vehicle:

$$\text{EMRDE} = \frac{1}{N}\sum_{i=1}^{N}\frac{|\hat{d}_i - d_i|}{d_i}, \tag{21}$$

$$\text{EMRAE} = \frac{1}{N}\sum_{i=1}^{N}\frac{|\hat{\theta}_i - \theta_i|}{\max(|\theta_i|, \epsilon)}, \tag{22}$$

where $d_i$ and $\theta_i$ are the ground truth distance and azimuth between $i$th participant and the ego vehicle, while $\hat{d}_i$ and $\hat{\theta}_i$ are the predicted values of VLMs, and $\epsilon$ is a small positive constant for numerical stability.

Object-centric metrics, which include Mean Relative Distance Error inter Object (OMRDE) and Mean Relative Angle Error inter Object (OMRAE), evaluate pairwise relationships among detected objects:

$$\text{OMRDE} = \frac{1}{N}\sum_{i=1}^{N}\frac{|\hat{d_{ij}} - d_{ij}|}{d_{ij}}, \tag{23}$$

$$\text{OMRAE} = \frac{1}{N}\sum_{i=1}^{N}\frac{|\hat{\theta_{ij}} - \theta_{ij}|}{\max(|\theta_{ij}|, \epsilon)}, \tag{24}$$

where $d_{ij}$ and $\theta_{ij}$ are the ground truth distance and azimuth between $i$th and $j$th participants, while $\hat{d_{ij}}$ and $\hat{\theta_{ij}}$ are the predicted values of VLMs, and $\epsilon$ is a small positive constant for numerical stability.

These indicators collectively assess spatial coherence and VLMs' ability to infer inter-object geometric dependencies.

## 3.3 Motion Planning

This module evaluates VLMs' dynamic reasoning ability, including both high-level meta-action prediction and low-level trajectory planning.

*3.3.1 Meta-action Prediction.* We define meta-actions based on acceleration and heading change trends derived from the annotated ground-truth 3-second future trajectory. Given a trajectory sequence $(x_t, y_t, z_t)$, the velocity $v_t$ and acceleration $a_t$ are computed as:

$$v_t = \frac{\sqrt{(x_{t+1} - x_t)^2 + (y_{t+1} - y_t)^2 + (z_{t+1} - z_t)^2}}{t_{t+1} - t_t}, \tag{25}$$

$$a_t = \frac{v_{t+1} - v_t}{t_{t+1} - t_t}, \tag{26}$$

and the heading and its change are defined as:

$$\theta_t = \text{atan2}(z_{t+1} - z_t, x_{t+1} - x_t), \tag{27}$$

$$\Delta\theta_t = \theta_{t+1} - \theta_t. \tag{28}$$

We define six meta-actions describing longitudinal and lateral behaviors, as summarized in Table 1.

**Table 1: Meta-action Classification Based on Acceleration and Change of Heading**

| Behavior Type | Acceleration ($a$) | Heading Change ($\Delta\theta$) |
|---|---|---|
| Accelerating | $a \geq +0.15m/s^2$ | – |
| Braking | $a \leq -0.15m/s^2$ | – |
| Constant Speed | otherwise | – |
| Left Turn | – | $\Delta\theta \geq +8°$ |
| Right Turn | – | $\Delta\theta \leq -8°$ |
| Go Straight | – | otherwise |

The Direction-Consistency Accuracy (DCS-Acc) is therefore introduced, which measures whether the VLM-predicted meta-action is consistent with the ground truth. DCS-Acc is formulated as follows:

$$\text{DCS-Acc} = \frac{1}{N}\sum_{i=1}^{N} I(A_i^{pred} = A_i^{gt}), \tag{29}$$

where $A_i^{pred}$ denotes meta-actions predicted by VLMs, and $A_i^{gt}$ denotes the ground truth.

We further use relative and quantitative metrics, such as Mean Relative Acceleration Error (MRE-Acc) and Angular Relative Error (ARE) defined as follows:

$$\text{MRE-Acc} = \frac{1}{N}\sum_{i=1}^{N}\frac{|a_i^{pred} - a_i^{gt}|}{|a_i^{gt}|}, \tag{30}$$

$$\text{ARE} = \frac{1}{N}\sum_{i=1}^{N}\frac{|\Delta\theta_i^{pred} - \Delta\theta_i^{gt}|}{\max(|\Delta\theta_i^{gt}|, \epsilon)}. \tag{31}$$

*3.3.2 Trajectory Planning.* Trajectory planning focuses on spatial accuracy and temporal stability of the VLM-predicted paths. Two standard metrics are employed: Average Displacement Error (ADE) and Final Displacement Error (FDE@T):

$$\text{ADE} = \frac{1}{N}\sum_{i=1}^{N}||\hat{p}_i - p_i||_2, \tag{32}$$

$$\text{FDE@T} = ||\hat{p_T} - p_T||_2. \tag{33}$$

ADE measures global spatial consistency, while FDE emphasizes endpoint accuracy. We report FDE@1s, FDE@2s, and FDE@3s to assess short-, mid-, and long-term prediction performance.

## 3.4 GPT-Score

We report GPT-Score, a semantic evaluation metric computed with GPT-4o. For each instance, GPT-4o receives the prompt, ground truth answer, and VLM prediction, and outputs a scalar in $[0, 1]$ measuring their semantic alignment (higher is better). For full-scene description and risk-reasoning samples, we additionally provide SPICE, Risk-Class-Acc, and the ground-truth risk level as auxiliary inputs. For other question types, GPT-4o only sees the prompt, ground truth, and prediction.

**Table 2: Normalization Parameters for Error-Based Metrics**

| Metrics / Parameters | $x_1$ | $x_2$ | $k$ |
|---|---|---|---|
| EMRDE | 0.1135 | 0.3856 | 9.1380 |
| EMRAE | 0.1052 | 0.4055 | 2.2210 |
| OMRDE | 0.1244 | 0.4252 | 8.0125 |
| OMRAE | 0.1155 | 0.4152 | 2.0542 |
| MRE-Acc | 0.7250 | 1.3056 | 0.0216 |
| ARE | 0.7588 | 1.3319 | 0.0125 |
| ADE | 2.2850 | 5.2278 | 7.2514 |
| FDE@1 | 1.3595 | 3.7750 | 6.7157 |
| FDE@2 | 1.3595 | 3.7750 | 6.7157 |
| FDE@3 | 1.3595 | 3.7750 | 6.7157 |

## 3.5 Evaluation Weighting Strategy

*3.5.1 Normalization.* All metrics in the benchmark are normalized into a standardized score range of $[0, 100]$ for non-error-based metrics, and $(0, 100]$ for error-based metrics. We distinguish non-error-based and error-based metrics, and separate normalization strategies tailored to their mathematical characteristics.

**Non-error-based Metrics:** Non-error-based metrics include SPICE, Risk-Class-Acc, Class-Acc, DCS-Acc, and GPT-Score. These metrics naturally take values in $[0, 1]$, and exhibit a simply "higher is better" monotonicity. For these metrics, we adopt a direct linear scaling to obtain a normalized score:

$$S_{non-error} = 100 \times M, \tag{34}$$

where $M$ denotes the raw metric value.

**Error-based Metrics:** Error-based metrics include EMRDE, EMRAE, OMRDE, OMRAE, MRE-Acc, ARE, ADE, and FDE@T. These metrics quantify deviations between predictions and ground truth, and therefore follow a simply "lower is better" monotonic behavior. We therefore use a piecewise normalization, mapping raw error $E$ to $(0, 100]$, which could be defined as:

$$S_{\text{error}}(E; x_1, x_2, k) = \begin{cases} 100, & E < x_1, \\ 100 - 40\,\dfrac{E - x_1}{x_2 - x_1}, & x_1 \le E < x_2, \\ 60\exp\big(-k(E - x_2)\big), & E \ge x_2, \end{cases} \tag{35}$$

where $x_1$ is the high precision threshold, $x_2$ is the acceptable error threshold, and $k > 0$ is an exponential decay factor.

We first estimate the normalization parameters for each error-based metric using the empirical error distributions from a representative set of baseline models. These parameters are then fixed and applied consistently to all subsequent model evaluations. Specifically, we set the high-precision threshold $x_1$ to the 15th percentile, and the acceptable error threshold $x_2$ to the 75th percentile of the raw errors. Thus, only the top 15% most accurate predictions receive a perfect score of 100, and the worst 25% predictions are mapped to the exponential penalty region.

This formulation provides the following desirable properties:

- **High-precision region ($E < x_1$):** Errors below a high-precision threshold are rewarded uniformly with a perfect score of 100, reflecting negligible deviation in practice.
- **Moderate-error region ($x_1 \le E < x_2$):** Scores decrease linearly from 100 to 60, enabling fine-grained comparison among VLMs with moderate errors.
- **High-error region ($E \ge x_2$):** Scores decay exponentially from 60 to 0, maintaining continuity while strongly penalizing large, potentially unsafe or physically infeasible errors more than small ones, in line with safety-critical evaluation in autonomous driving.

Table 2 lists the normalization parameters for all error-based metrics, showing how each maps into the unified scoring space.

**Table 3: Weighting structure of the evaluation framework of ScenePilot-Bench benchmark**

| Module | Module Weight | Metric | Metric Weight |
|---|---|---|---|
| Scene Understanding | 0.15 | SPICE | 0.70 |
| | | Risk-Class-Acc | 0.30 |
| Spatial Perception | 0.35 | Class-Acc | 0.20 |
| | | EMRDE | 0.30 |
| | | EMRAE | 0.20 |
| | | OMRDE | 0.20 |
| | | OMRAE | 0.10 |
| Motion Planning | 0.40 | DCS-Acc | 0.20 |
| | | MRE-Acc | 0.10 |
| | | ARE | 0.10 |
| | | FDE@1 | 0.10 |
| | | FDE@2 | 0.10 |
| | | FDE@3 | 0.20 |
| | | ADE | 0.20 |
| GPT-Score | 0.10 | GPT Alignment Score | 1.00 |

*3.5.2 The Overall Weighting Strategy.* As shown in Table 3, the evaluation framework adopts a structured weighting strategy that reflects the functional roles and safety relevance of different system capabilities.

Within each module, individual metrics are further weighted based on their operational significance. Within scene understanding, we prioritize semantic robustness measured by SPICE, while still allocating substantial weight to risk classification due to its safety relevance. Spatial Perception emphasizes ego-relative distance and angle errors, which directly affect collision risk, while inter-object metrics capture interaction reasoning complexity. Motion Planning integrates both dynamic correctness and multi-horizon endpoint accuracy (FDE@1/2/3), reflecting the increasing difficulty and importance of long-term trajectory prediction, complemented by ADE for overall trajectory consistency.

## 3.6 Geographic Generalization

The robustness against domain shifts and geographic variations is crucial for real-world deployment. To comprehensively evaluate VLMs' robustness and adaptability across unseen geographic regions and camera calibrations, we design a systematic generalization assessment protocol comprising two cross-domain settings. No new metrics are introduced in this section.

**LOCO (Leave-One-Country-Out) Evaluation.** We adopt a LOCO strategy to assess geographic robustness. To evaluate cross-country generalization, VLMs are trained solely on data from China and tested on held-out datasets from four regions: Europe (EU), Japan/Korea (JP/KR), the United States (US), and other countries (OTH).

**Right-to-Left Adaptation.** Driving environment differs significantly between left-hand and right-hand traffic systems. To evaluate VLMs' adaptability to this fundamental shift, we introduce a cross-traffic evaluation:

- R→L: Train exclusively on right-hand traffic countries (e.g., China, US), test on left-hand traffic countries (e.g., Japan, UK).

Performance drop relative to an in-domain baseline measures resilience to new traffic rules, which means that smaller drop reflects better generalization.

## 4 Implementation Details of Experiments

In the comparative experiment of various VLMs on the ScenePilot-Bench benchmark, the original dataset is partitioned into training, validation, and test sets respectively. The training set is used for VLM fine-tuning, the validation set for hyperparameter tuning and model selection, and the test set for final performance evaluation. The split follows a principle of multidimensional conditional uniformity, jointly accounting for weather conditions, road types, lane configurations, intersection attributes, risk levels, traffic density, and geographic regions to ensure balanced representation across all subsets.

In constructing the VQA dataset, samples for the training and test sets are independently drawn from their respective video splits to rigorously prevent data leakage. Each VQA sample is associated with a unique video clip, ensuring no overlap between training and test samples. This design guarantees that model evaluation reflects true generalization to unseen scenarios. Although we annotate nearly 400 million VQA samples for the entire dataset, such a volume is unnecessary for the current fine-tuning of VLMs. In the comparative experiment, the training and test sets contain 200,000 and 100,000 VQA samples, respectively, which are sufficient to demonstrate the effectiveness of our proposed method. The full annotated dataset could be used in the future to train the foundational capabilities of VLMs.

In this work, we fine-tune VLMs using Parameter-Efficient Fine-Tuning (PEFT) strategies within the Unsloth optimization framework. Experiments were conducted on four NVIDIA A800 GPUs employing a Distributed Data Parallel (DDP) strategy. To balance memory efficiency with numerical stability, we utilized Bfloat16 precision throughout the process, in conjunction with the 8-bit AdamW optimizer.

## 5 Cross-Dataset Validation on Annotation

### 5.1 Ego Trajectory Annotation

As shown in Fig. 2, 3, and 4, we present examples of accurate and failed ego trajectory annotations in the nuScenes dataset. The accurate cases demonstrate that the annotated trajectories closely follow the actual movement of the ego vehicle, while the failed case shows significant deviation from the true path, indicating potential issues in the annotation process.

The accurate annotations show that the trajectory points are well-aligned with the road and follow a smooth path consistent with typical driving behavior. In contrast, the failed annotation exhibits erratic points that do not align with the road, suggesting errors in the camera calibration or depth estimation process.

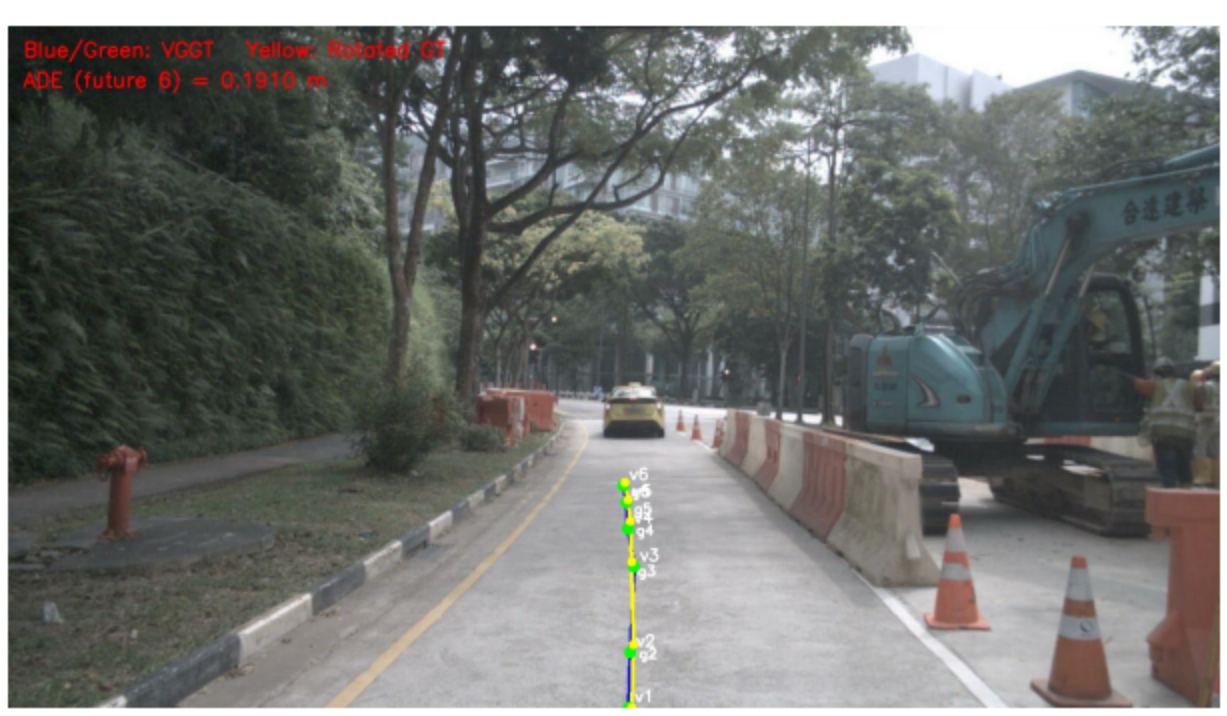

**Figure 2: Accurate annotation of trajectories in a driving scene of nuScenes.**

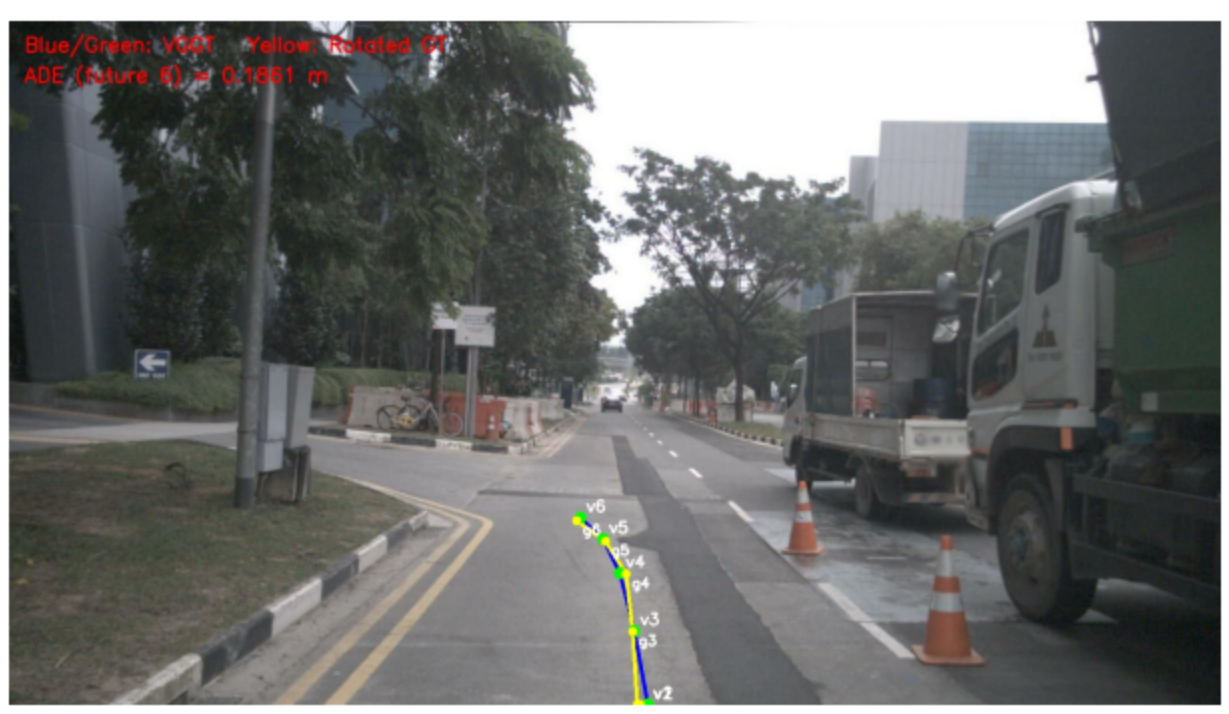

**Figure 3: Accurate annotation of trajectories in a driving scene of nuScenes.**

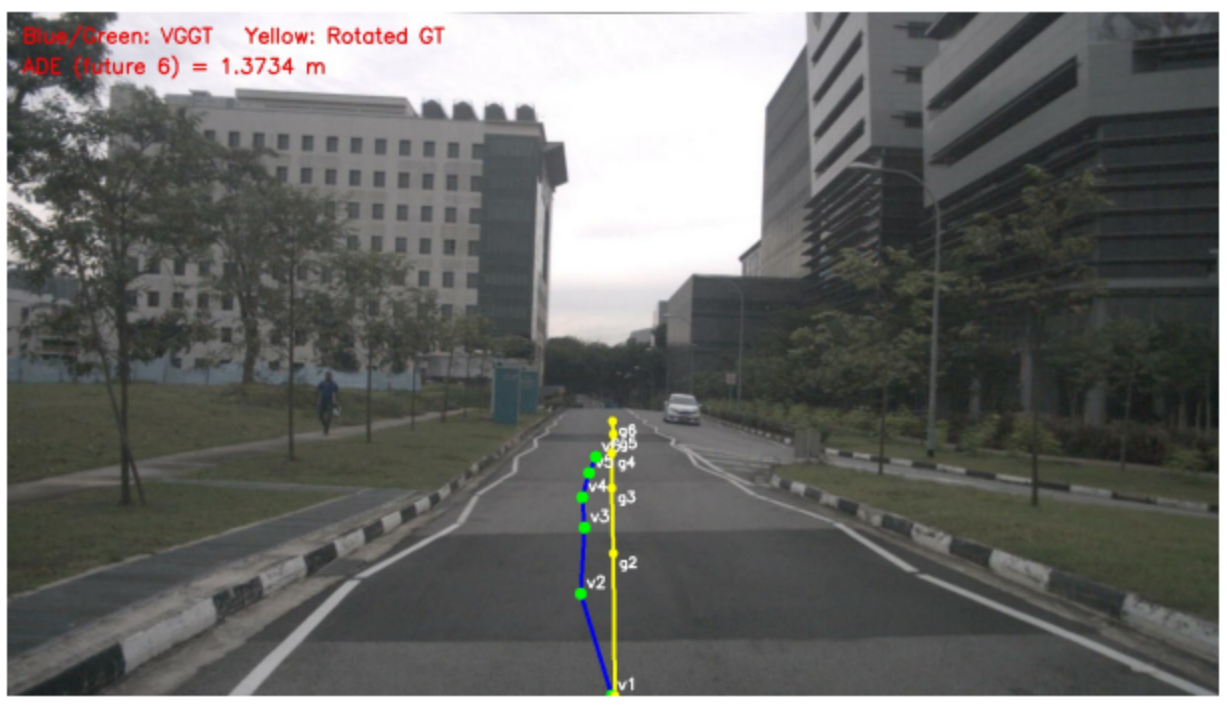

**Figure 4: Failed annotation of trajectories in a driving scene of nuScenes.**

To improve the statistical consistency and physical validity of the trajectory annotations, we apply a rule-based multi-criteria filtering module to systematically screen raw trajectory sequences and remove abnormal samples. Given a trajectory sample, we first concatenate the historical and future segments into a complete temporal sequence in the $x$–$z$ plane,

$$\mathcal{T} = \{(x_t, z_t)\}_{t=1}^{N}. \tag{36}$$

Based on this full sequence, we compute a set of basic motion descriptors, including the total path length, speed statistics, and heading variation, and then evaluate trajectory quality from three complementary aspects: motion magnitude, geometric shape, and dynamical consistency.

For motion magnitude, we identify low-information trajectories by jointly examining the average speed, maximum speed, and total path length. Samples with extremely small motion magnitude are treated as invalid, since they usually correspond to static or near-static cases and contribute limited supervisory value for trajectory prediction. Let

$$L = \sum_{t=2}^{N} \|\mathbf{p}_t - \mathbf{p}_{t-1}\|_2, \qquad \mathbf{p}_t = (x_t, z_t), \tag{37}$$

denote the path length, and let $v_t$ denote the instantaneous speed computed from consecutive points. We then use the average speed $\bar{v}$, the maximum speed $v_{\max}$, and $L$ as joint criteria for low-motion filtering.

For geometric shape, we detect implausible lateral drift and abnormal high-curvature patterns from local displacement and heading changes. Specifically, consecutive displacements are analyzed in the $x$–$z$ plane to identify samples with excessive lateral deviation, and heading-angle variation is further used to detect trajectories with unrealistic turning behavior. At the same time, a straight-motion prior is introduced to avoid over-penalizing samples that follow approximately linear motion, thereby reducing false rejection of valid trajectories.

For dynamical consistency, we evaluate whether the trajectory exhibits physically smooth motion evolution. We construct an acceleration sequence from three-point finite differences and further inspect the norm change between adjacent accelerations. If the acceleration variation changes abruptly, the corresponding trajectory is regarded as dynamically inconsistent, which usually indicates discontinuous motion patterns or reconstruction artifacts. Formally, with a fixed sampling interval $\Delta t$, the acceleration can be approximated as

$$\mathbf{a}_t = \frac{\mathbf{p}_{t+1} - 2\mathbf{p}_t + \mathbf{p}_{t-1}}{(\Delta t)^2}, \tag{38}$$

and abrupt dynamical changes are detected from the variation of $\|\mathbf{a}_t - \mathbf{a}_{t-1}\|_2$.

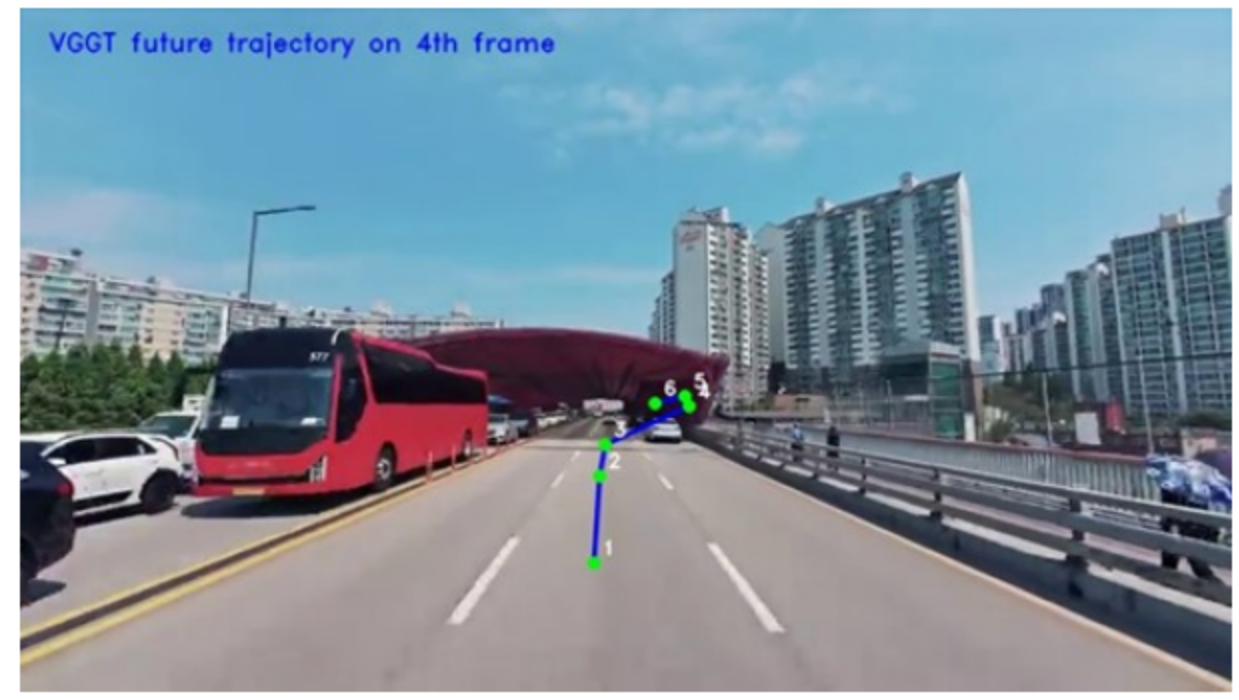

**Figure 5: Failed annotation of trajectories in a driving scene of Korea, is removed from the dataset.**

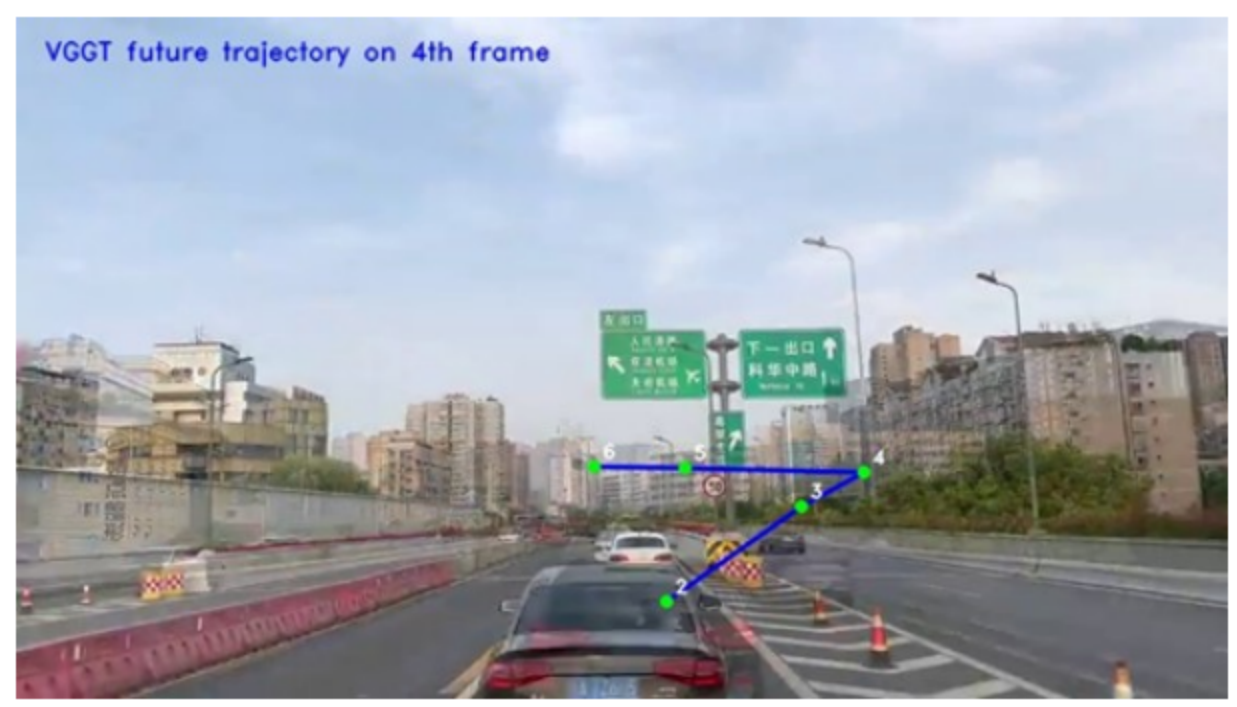

**Figure 6: Failed annotation of trajectories in a driving scene of China, is removed from the dataset.**

Once any abnormality criterion is triggered, the corresponding sample is marked as invalid and its trajectory sequence is removed from the dataset. However, the associated metadata are retained for dataset-level distribution statistics and subsequent error analysis. In addition, we record both the anomaly type and the sample index for each rejected trajectory, which supports quantitative threshold calibration and later inspection of failure modes. By jointly constraining motion magnitude, geometric structure, and dynamical consistency, this cleaning strategy effectively reduces the proportion of low-information and physically implausible trajectories, thereby improving the overall reliability of the dataset. Trajectories, such as those shown in Fig. 5 and 6, are filtered out by this module due to their significant deviation from typical driving patterns, which could be caused by issues in the annotation process or sensor errors.

## 5.2 Spatial Perception Annotation

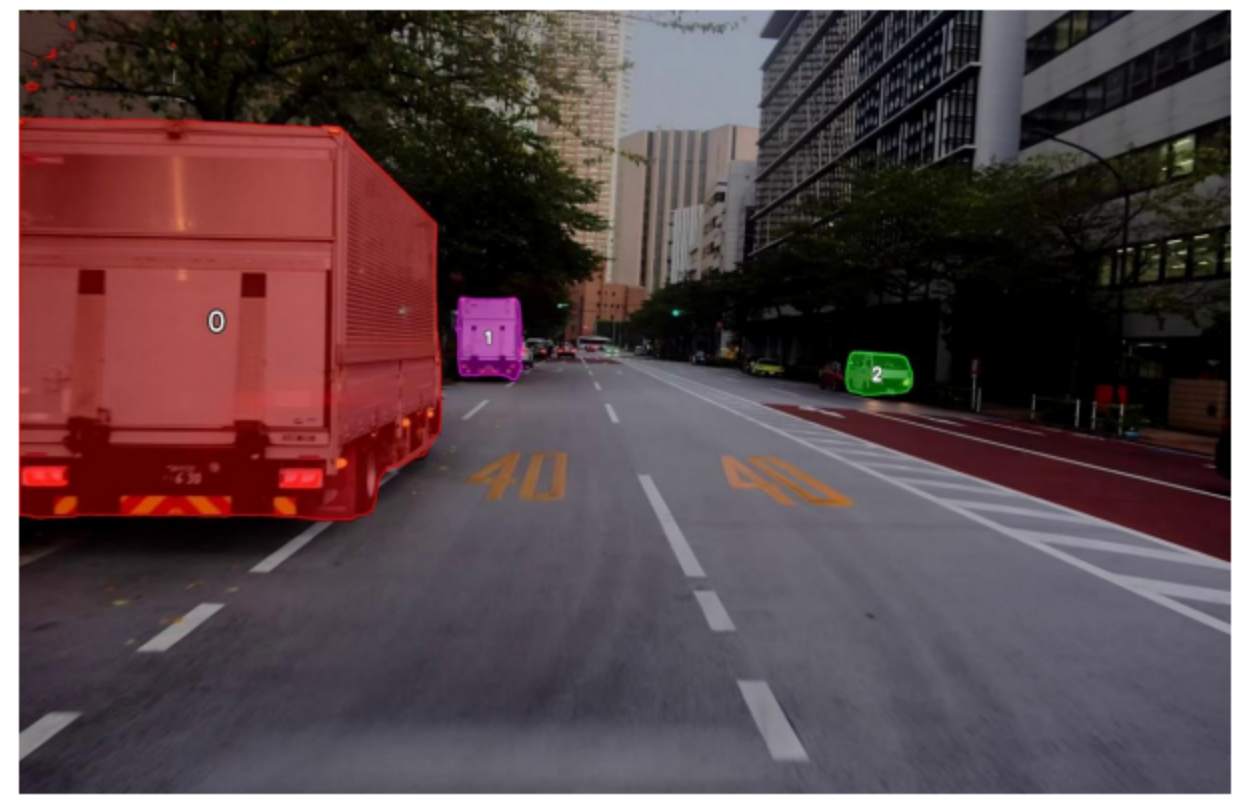

**Figure 7: Accurate annotation of spatial-perception labels.**

As shown in Fig. 7 and 8, we present examples of accurate and failed spatial perception annotations in the STRIDE-QA dataset.

In Fig. 7, which is the accurate case, the spatial perception labels correctly identify the positions of the traffic participants. For example, the ground truth of the distance between the truck in region [2] and the ego vehicle is 43.47 meters, and the annotated label is 45.61 meters, with the error within 5%. The azimuth label is also accurate, with a ground truth of -15.00 degrees and an annotation of -14.68 degrees, which is within a reasonable error margin.

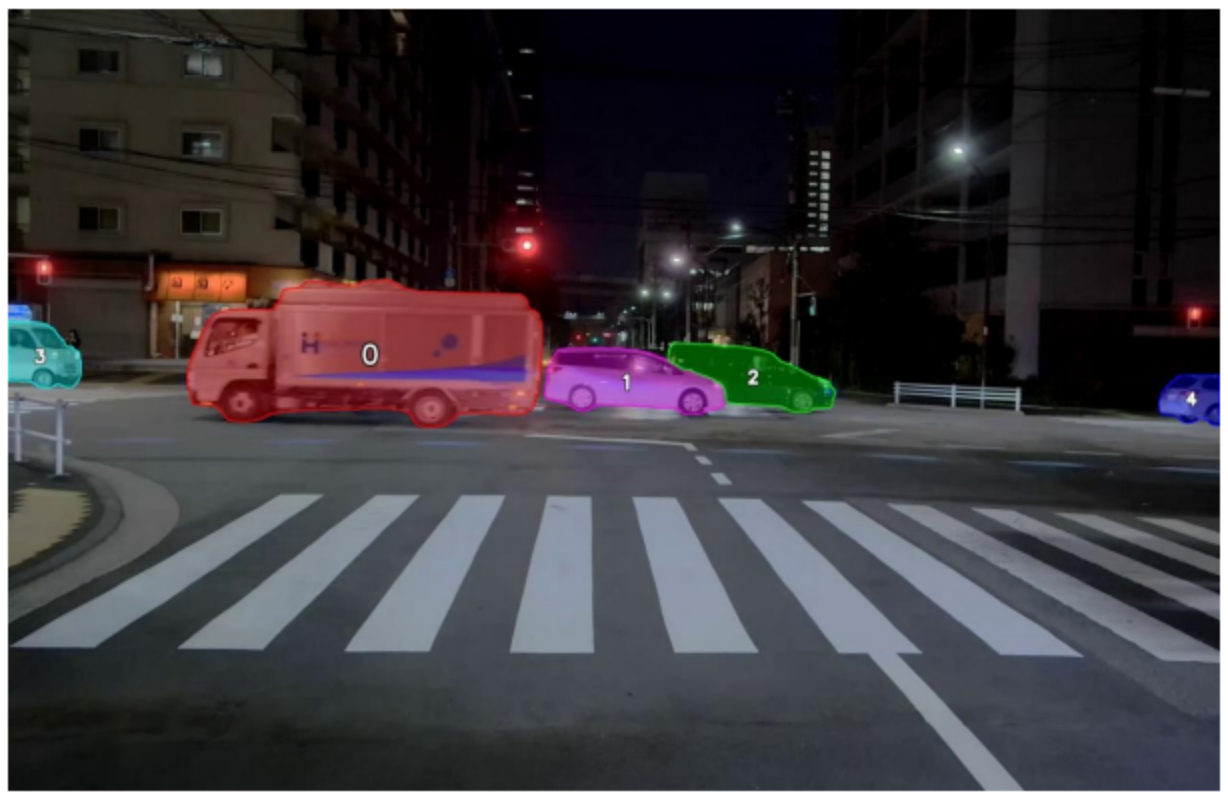

**Figure 8: Failed annotation of spatial-perception labels.**

In Fig. 8, which is the failed case, the spatial perception labels show deviations from the ground truth. For instance, the distance between the car in region [1] and the ego vehicle has a ground truth of 24.77 meters, but the annotated label is 30.76 meters, which is a large error of approximately 24.2%. The azimuth label also shows a significant error, with a ground truth of -3.51 degrees and an annotation of -0.76 degrees.

But the number of failed cases is relatively small compared to the accurate cases, and the overall annotation quality is high. We choose 500 samples from the STRIDE-QA dataset for manual inspection, and define a sample as failed if either the distance or azimuth error exceeds 8%. The failure rate is approximately 3%, which indicates that the majority of annotations are accurate and reliable for training and evaluation purposes.